\documentclass[twocolumn,pre]{revtex4}   

\usepackage{dcolumn}
\usepackage{amsmath}

\usepackage{graphicx}

\newcommand{\e}{\mathrm{e}}
\newcommand{\dd}{\mathrm{d}}
\newcommand{\etal}{{\it{}et~al.}}

\begin{document}

\title{Ranking with multiple types of pairwise comparisons}
\author{M. E. J. Newman}
\affiliation{Center for the Study of Complex Systems, University of Michigan, Ann Arbor, MI 48109, USA}

\begin{abstract}
  The task of ranking individuals or teams, based on a set of comparisons between pairs, arises in various contexts, including sporting competitions and the analysis of dominance hierarchies among animals and humans.  Given data on which competitors beat which others, the challenge is to rank the competitors from best to worst.  Here we study the problem of computing rankings when there are multiple, potentially conflicting modes of comparison, such as multiple types of dominance behaviors among animals.  We assume that we do not know \textit{a priori} what information each behavior conveys about the ranking, or even whether they convey any information at all.  Nonetheless we show that it is possible to compute a ranking in this situation and present a fast method for doing so, based on a combination of an expectation-maximization algorithm and a modified Bradley-Terry model.  We give a selection of example applications to both animal and human competition.
\end{abstract}

\maketitle

\section{Introduction}
Suppose a set of individuals or teams compete with one another in pairwise fashion, as in a sports league or chess tournament.  The outcome of each game is recorded and your task is to rank the players or teams from best to worst based on those outcomes.  Similar ranking problems also arise in paired comparison tests in consumer research and behavioral psychology, and in dominance hierarchies in animals and humans, where individuals compete in pairwise fashion to establish dominance and the challenge is to estimate the hierarchy of individuals from their observed interactions.

Ranking problems of this kind have been widely studied for over a century~\cite{Thurstone27,Zermelo29,BT52,David88,DF76,Cattelan12}.  They can be non-trivial because the outcomes of competitions are not always consistent---better players sometimes lose and worse players win---but excellent methods exist for computing rankings nonetheless.  Here we study the more complex ``multimodal'' problem of computing rankings when there is more than one type or mode of competition between individuals.  Many animals, for instance, compete in a range of different ways.  Mountain goats attack opponents with their horns but also employ a type of rushing attack~\cite{Cote00}.  Spotted hyenas also use a rushing attack, or they may bite their opponents~\cite{Frank86}.  In traditional ranking analyses one treats these modes as equivalent, but different modes may contribute more or less to establishing dominance and by exploiting this variation one can perform a more fine-grained analysis that reveals things not readily accessible via standard methods.

Ranking analyses are frequently couched in the language of wins and losses, but there are many cases where this framing does not apply.  In observations of interactions among elementary school children, for instance, there is often a well established hierarchy but many interactions are not overtly competitive.  One child directing the actions of another, for instance, is commonly observed and non-competitive but is nonetheless believed to be indicative of dominance~\cite{Golemiec16}.  In cases like this merely performing the action is the equivalent of a win.  Similar situations also arise in animal behavior, where dominance displays without physical competition are often preferred by both parties to avoid injury.  American bison, for example, do fight one another, but they also employ a range of non-contact threats including a ``nod threat'' and a ``broadside threat''~\cite{Lott79}.  Dominance is signaled merely by performing the action.

On the other hand, not all actions necessarily indicate dominance: some indicate subordination.  In the case of schoolchildren, for instance, actions like directing behavior indicate dominance, but actions like following, copying, or watching typically indicate subordination~\cite{Golemiec16}.  Both dominant and subordinate actions may be useful for inferring hierarchy, but we need to know which is which in order to analyze them appropriately.

In the work presented here, we explicitly do not assume that we know the valence of the various types of action performed in a particular setting.  However, it turns out, as we will show, that one can infer it automatically from observational data, and once we know the nature of the interactions we can use this information to make an estimate of the complete ranking of individuals or teams, weighting each interaction type appropriately.  Interactions that are strongly correlated with dominance carry heavy weight; those that are strongly correlated with subordination do so as well, but with an opposite sign.  And interactions that are only weakly correlated are given low weight.  In this way, we can construct a ranking that correctly allows for the different ways in which individuals interact.

A number of previous authors have studied related problems.  Our multimodal ranking is in some ways reminiscent of the ``multivariate'' comparisons studied by Davidson and Bradley and others~\cite{DB69,Bockenholt88,DFHK06}.  These authors consider problems of consumer choice in which items or products are rated by consumers on several criteria, such as a car rated on price, size, fuel economy, and so forth.  A crucial difference between these studies and ours is that our goal is to infer only a single, one-dimensional ranking of items or individuals, whereas the multivariate approach produces multiple rankings, one for each criterion.  While the multivariate approach is appropriate in settings such as consumer choice, the approach of the present paper is more appropriate for applications to dominance hierarchies, sports, and other competitions where we either desire or believe axiomatically that the individuals or teams exist on a single rank scale.  Perhaps a closer parallel to our approach is to be found in the literature on rank aggregation methods~\cite{Lin10}.  These are general methods for taking multiple, possibly conflicting rankings of the same set of items or individuals and combining them to generate a single consensus ranking.  (Note that the phrase ``rank aggregation'' is also sometimes used to describe the unimodal ranking problem~\cite{Lin10,RA14}, so one must be careful when reading this literature.)  Rank aggregation is used for instance to combine rankings of web pages for output by web search engines~\cite{DKNS01}.  Although they are based on different principles to our approach, such as numerical minimization of the distance between rankings, rank aggregation methods could be applied to the types of problems we consider by generating separate rankings for each criterion or mode and then aggregating them.  A related approach has been taken by Pritikin~\cite{Pritikin20}, who also constructs different ranking scales for different criteria, but then performs a factor analysis to construct a consensus measure of merit.  Like the multivariate approach, these methods contrast philosophically with our assumption that there only exists a single ranking, but the outputs could give similar results in some cases.

\section{The model}
\label{sec:method}
Suppose we have a population of $N$ individuals or teams labeled by $u = 1\ldots N$ and a sequence of contests or interactions between them.  Let there be $M$ interactions in total denoted by $r = 1\ldots M$, let $u_r$ be the individual who wins the $r$th interaction (in cases where there is a winner) or who instigates the interaction (in cases where there is no winner).  Similarly let $v_r$ be the loser or recipient of the $r$th interaction.  For simplicity we assume that there are no ties or draws, although the approach we describe could be generalized to the case of ties using standard methods~\cite{RK67,Davidson70}.

For each interaction we assume that there is a dominant individual and a subordinate individual.  Bear in mind, however, that the dominant individual need not be the winner or instigator of the interaction: some games may be won by the weaker player and some actions may be instigated by a subordinate individual.  Moreover, we explicitly allow for the possibility that the dominant individual in a pair may change from one interaction to another.  In some pairs one individual may always be dominant, but in others the roles may vary.  We define a stance variable~$\sigma_r$, which is 1 if the winner~$u_r$ is the dominant individual during interaction~$r$ and 0 if $v_r$~is dominant.  (To simplify the discussion we will henceforth refer to $u_r$ as the ``winner,'' but this should be taken to include the instigator in cases where merely performing an action is equivalent to a win, and similarly for losses.)

We assume the stance is independent for each interaction and model it using a standard Bradley-Terry model of dominance~\cite{Zermelo29,BT52,David88}.  We define a score or ranking~$s_u$ for each individual~$u$ and we assume that the probability~$p_{uv}$ of individual $u$ dominating over individual~$v$ on any particular interaction is a function $f(s_u-s_v)$ of the difference of their scores.  In the Bradley-Terry model this function is chosen to be the logistic function $f(s) = 1/(1+\e^{-s})$, which means that
\begin{equation}
p_{uv} = {\e^{s_u}\over\e^{s_u}+\e^{s_v}}.
\end{equation}
These probabilities are invariant under a uniform shift of all the scores~$s_u$ by an additive constant, so one commonly introduces some normalization or standardization to fix the origin of the score scale.  Here we do this by fixing the average value of the scores to be zero, so that $s=0$ indicates an average individual who is equally likely to be dominant or subordinate.

For convenience, one also often introduces the shorthand $\lambda_u = \e^{s_u}$, so that
\begin{equation}
p_{uv} = {\lambda_u\over\lambda_u+\lambda_v},
\label{eq:puv}
\end{equation}
and we will do that here.  Following Zermelo~\cite{Zermelo29} we refer to $\lambda_u$ as a strength parameter or simply a strength.  Note that for the average individual with $s=0$ we have $\lambda=\e^0=1$, so the probability $p_0$ of an individual with strength parameter~$\lambda$ dominating against the average individual is
\begin{equation}
p_0 = {\lambda\over\lambda+1},
\label{eq:p0}
\end{equation}
and hence $\lambda = p_0/(1-p_0)$.  In other words, $\lambda$ has a simple interpretation: it is the odds of dominating against the average individual.

The winner of an interaction need not, as we have said, be the dominant individual.  Sometimes the winner is the subordinate individual, and the frequency with which this happens may depend on the type of the interaction.  Let there be $T$ interaction types, labeled by $t=1\ldots T$.  We define another parameter~$q_t$, which we call the \textit{valence probability}, equal to the probability that the currently dominant individual wins an interaction of type~$t$ (and $1-q_t$ is the probability that the subordinate individual wins).

Note the distinction here between the \emph{dominant individual} and the \emph{winner}.  In the conventional theory of unimodal paired comparisons no distinction is made between these: the probability~$p_{uv}$ of Eq.~\eqref{eq:puv} directly gives the probability of winning and there is no equivalent of the valence probability~$q_t$.  In the multimodal case, however, this approach is insufficient because it fails to separate the properties of the individuals from the properties of the different interaction types.  The model proposed here allows for the possibility that different interaction types may have different probabilities of being instigated or won, even by the same individual.

Given these definitions, we can now write down an expression for the likelihood of occurrence of a particular sequence of wins and losses.  Consider an interaction~$r$ with winner~$u$, loser~$v$, and type~$t$.  The probability of the stance variable taking value~$\sigma_r=1$ is equal to $p_{uv}$ and the probability that $\sigma_r=0$ is $1-p_{uv}$.  In general,
\begin{equation}
P(\sigma_r|\lambda_u,\lambda_v) =  p_{uv}^{\sigma_r} (1-p_{uv})^{1-\sigma_r}
  = {\lambda_u^{\sigma_r}\lambda_v^{1-\sigma_r}\over \lambda_u+\lambda_v}.
\end{equation}

Given the value of~$\sigma_r$ we can calculate the probability that individual~$u$ did indeed win: if $\sigma_r=1$, so that $u$ is dominant, then the probability is~$q_t$; if $\sigma_r=0$, so that $u$ is subordinate, then the probability is~$1-q_t$.  Denoting the observation data for interaction~$r$ by~$x_r$, these probabilities can be compactly combined as
\begin{equation}
P(x_r|\sigma_r,q_t) = q_t^{\sigma_r} (1-q_t)^{1-\sigma_r}.
\label{eq:win}
\end{equation}
Then the probability that the stance is~$\sigma_r$ and $u$ wins is
\begin{align}
P(x_r,\sigma_r|\lambda_u,\lambda_v,q_t) &= P(x_r|\sigma_r,q_t)
      P(\sigma_r|\lambda_u,\lambda_v) \nonumber\\
  &= (p_{uv}q_t)^{\sigma_r} \bigl[(1-p_{uv})(1-q_t)\bigr]^{1-\sigma_r} \nonumber\\
  &= {(\lambda_uq_t)^{\sigma_r} [\lambda_v(1-q_t)]^{1-\sigma_r}\over
      \lambda_u+\lambda_v}.
\label{eq:onep}
\end{align}

Finally, under the assumption that interactions~$r$ are independent of one another, the likelihood of the entire set of observations is the product of this expression over all~$r$, which gives
\begin{align}
P(x,\sigma|\lambda,q) = \prod_{r=1}^M
  {(\lambda_{u_r}q_{t_r})^{\sigma_r} [\lambda_{v_r}(1-q_{t_r})]^{1-\sigma_r}\over
    \lambda_{u_r}+\lambda_{v_r}},
\label{eq:likelihood}
\end{align}
where the unsubscripted variables $x$, $\sigma$, $\lambda$, and~$q$ indicate the complete sets of data and parameters.

Our goal is to use Eq.~\eqref{eq:likelihood} to estimate the values of~$\lambda_u$ and~$q_t$ for all individuals~$u$ and all interaction types~$t$.

\section{EM algorithm}
One might imagine that the next step in the calculation would be to sum over the variables~$\sigma_r$ in Eq.~\eqref{eq:likelihood}, which can be done easily to give an expression for~$P(x|\lambda,q)$ thus:
\begin{equation}
P(x|\lambda,q) = \sum_\sigma P(x,\sigma|\lambda,q)
  = \prod_{r=1}^M  {\lambda_{u_r}q_{t_r} + \lambda_{v_r}(1-q_{t_r})\over
    \lambda_{u_r}+\lambda_{v_r}}.
\label{eq:PD}
\end{equation}
Then we could maximize this expression with respect to $\lambda$ and~$q$ to find maximum-likelihood estimates of the parameters, or alternatively maximize its logarithm, which is equivalent and usually simpler.  Though the derivatives can be done, however, they yield a complicated set of implicit equations with no easy solution.  Instead, therefore, we adopt a different approach, making use of an expectation-maximization (EM) algorithm.

Instead of maximizing $\log P(x|\lambda,q)$ directly, we apply Jensen's inequality, which says for any set of non-negative quantities $z_i$ and weights~$\pi_i$ satisfying~$\sum_i \pi_i=1$ that $\log \sum_i z_i \ge \sum_i \pi_i \log(z_i/\pi_i)$.  Applied to the present case, this gives
\begin{align}
\log P(x|\lambda,q) &= \log \sum_\sigma P(x,\sigma|\lambda,q) \nonumber\\
  &\ge \sum_\sigma \pi(\sigma) \log {P(x,\sigma|\lambda,q)\over\pi(\sigma)},
\label{eq:jensen}
\end{align}
where $\pi(\sigma)$ is any probability distribution over the set~$\sigma$ of stance variables satisfying $\sum_\sigma \pi(\sigma) = 1$.  The exact equality is achieved---and hence the right-hand side of~\eqref{eq:jensen} maximized---when
\begin{align}
\pi(\sigma) &= {P(x,\sigma|\lambda,q)\over\sum_\sigma P(x,\sigma|\lambda,q)}
     \nonumber\\
  &= {\prod_r \bigl( \lambda_{u_r} q_{t_r} \bigr)^{\sigma_r}
      \bigl[ \lambda_{v_r}(1-q_{t_r}) \bigr]^{1-\sigma_r}\over
      \prod_r \bigl[ \lambda_{u_r} q_{t_r} + \lambda_{v_r}(1-q_{t_r}) \bigr]}
      \nonumber\\
  &= \prod_r \pi_r^{\sigma_r} (1-\pi_r)^{1-\sigma_r},
\label{eq:estep1}
\end{align}
where
\begin{equation}
\pi_r = {\lambda_{u_r} q_{t_r}\over
         \lambda_{u_r} q_{t_r} + \lambda_{v_r}(1-q_{t_r})}
\label{eq:estep2}
\end{equation}
and we have used Eqs.~\eqref{eq:likelihood} and~\eqref{eq:PD}.  The quantity~$\pi_r$ can be interpreted as the posterior probability that $u_r$ is the dominant participant in interaction~$r$.

Since the choice~\eqref{eq:estep1} maximizes the right-hand side of~\eqref{eq:jensen} and simultaneously makes the two sides equal, a further maximization with respect to $\lambda$ and~$q$ will then achieve our goal of finding the maximum-likelihood values of these parameters.  Equivalently, a double maximization of the right-hand side with respect to both~$\pi(\sigma)$ and the parameters~$\lambda,q$ will achieve the same goal.  In an EM algorithm we perform this double maximization by alternately and repeatedly maximizing with respect to~$\pi(\sigma)$ and with respect to~$\lambda,q$ until convergence is reached.  The first we do using Eqs.~\eqref{eq:estep1} and~\eqref{eq:estep2} and the second by differentiation as follows.

\begin{widetext}
Substituting from Eqs.~\eqref{eq:likelihood} and~\eqref{eq:estep1} into~\eqref{eq:jensen} and neglecting terms that do not depend on~$\lambda,q$, we find that
\begin{align}
\sum_\sigma \pi(\sigma) &\log P(x,\sigma|\lambda,q)
   = \sum_\sigma \biggl[ \prod_r \pi_r^{\sigma_r} (1-\pi_r)^{1-\sigma_r} \biggr]
     \sum_r \Bigl[ \sigma_r \log ( \lambda_{u_r} q_{t_r} ) +
     (1-\sigma_r) \log\bigl[ \lambda_{v_r}(1-q_{t_r}) \bigr]
     - \log(\lambda_{u_r}+\lambda_{v_r}) \Bigr] \nonumber\\
  &= \sum_r \Bigl[ \pi_r \log ( \lambda_{u_r} q_{t_r} ) +
     (1-\pi_r) \log\bigl[ \lambda_{v_r} (1-q_{t_r}) \bigr]
     - \log(\lambda_{u_r}+\lambda_{v_r}) \Bigr] \nonumber\\
  &= \sum_r \sum_{uv} \delta_{u_ru}\delta_{v_rv}
     \bigl[ \pi_r \log \lambda_u + (1-\pi_r) \log \lambda_v
     - \log(\lambda_u+\lambda_v) \bigr]
     + \sum_r \sum_t \delta_{t_rt} \bigl[ \pi_r \log q_t
     + (1-\pi_r) \log(1-q_t) \bigr],
\label{eq:biglikelihood}
\end{align}
where $\delta_{ab}$ is the Kronecker delta.
\end{widetext}

Now we differentiate this expression to calculate the maximum-likelihood estimates of the parameters.  Differentiating with respect to~$q_t$ gives the straightforward result
\begin{equation}
q_t = {\sum_r \delta_{t_rt} \pi_r\over\sum_r \delta_{t_rt}}.
\label{eq:mstepq}
\end{equation}
Calculating the $\lambda$ parameters is a little more complicated.  Differentiating~\eqref{eq:biglikelihood} with respect to~$\lambda_i$ and setting the result to zero, we get
\begin{equation}
\sum_r {\delta_{u_ri} \pi_r + \delta_{v_ri}(1-\pi_r)\over\lambda_i}
  - \sum_r \biggl[ {\delta_{u_ri}\over\lambda_i+\lambda_{v_r}}
  + {\delta_{v_ri}\over\lambda_{u_r}+\lambda_i} \biggr] = 0,
\label{eq:mstepzero}
\end{equation}
which can be rearranged into the form
\begin{equation}
\lambda_i = {\sum_r \bigl[ \pi_r \delta_{u_ri} + (1-\pi_r) \delta_{v_ri} \bigr]
             \over \sum_u A_{iu}/(\lambda_i+\lambda_u)},
\label{eq:msteplambda}
\end{equation}
where
\begin{equation}
A_{ij} = \sum_r \bigl[ \delta_{u_ri}\delta_{v_rj} + \delta_{u_rj}\delta_{v_ri}
        \bigr]
\end{equation}
is the total number of times that $i$ and $j$ interact.  Equation~\eqref{eq:msteplambda} can be solved for~$\lambda_i$ by simple iteration starting from any convenient set of initial values $\lambda_i>0$.  This iterative procedure can be thought of as a variation on the well-known algorithm of Zermelo~\cite{Zermelo29} for the traditional Bradley-Terry model with just a single mode of interaction.  (For small problems with only a few individuals or teams it may be faster to solve~\eqref{eq:mstepzero} directly by Newton's method or Fisher scoring, but the speed advantage is relatively small and for larger instances the simple iterative approach is faster because it does not require a matrix inversion.  On balance, we recommend the iterative approach in most cases.)

The full algorithm for computing a ranking now consists of the following steps:
\begin{enumerate}
\item Choose initial values for the parameters~$\lambda_u$ and $q_t$ for all individuals~$u$ and interaction types~$t$.  These values can for instance be chosen at random.
\item Compute the quantities~$\pi_r$ from Eq.~\eqref{eq:estep2}.
\item Compute new estimates of the valence probabilities~$q_t$ from Eq.~\eqref{eq:mstepq}.
\item Compute new estimates of the strengths~$\lambda_u$ by iterating Eq.~\eqref{eq:msteplambda} to convergence.
\item Normalize the $\lambda_u$ so that the average individual has rank score zero, which is equivalent to dividing by the geometric mean of the $\lambda_u$ thus:
\begin{equation}
\lambda_u \leftarrow {\lambda_u\over\bigl( \prod_{v=1}^N \lambda_v \bigr)^{1/N}}.
\end{equation}
\item Repeat from step~2 until overall convergence is achieved.
\end{enumerate}

The end result is a complete set of strengths~$\lambda_u$ that specify the ranking of the individuals---higher~$\lambda_u$ implies higher ranking.  If we wish, we can convert these parameters back to the original scores $s_u = \log \lambda_u$, which are more symmetric and arguably easier to interpret.  A~minor technical issue is that the likelihood of Eq.~\eqref{eq:PD} is invariant under the change $\lambda_u \to 1/\lambda_u$, $q_t\to1-q_t$ for all $u$ and~$t$, meaning that, depending on the initial conditions, the final ranking may end up being upside down, so that the most dominant individuals have the lowest scores and the most subordinate ones have the highest.  If this happens one need merely invert all the values of the $\lambda_u$ to set them the right way up.

\section{Prior on the strength parameters}
\label{sec:map}
The procedure described in the previous section amounts to a complete algorithm for calculating the maximum-likelihood values of all the parameters in our model.  In some cases, however, this method gives poor results, for well understood reasons.  Maximum likelihood estimates of Bradley-Terry style models can perform poorly because they give undue weight to very large and very small values of the scores~$s_u$.  This is because the maximum-likelihood approach effectively assumes an (improper) uniform prior on~$s_u$, but $s_u$ has infinite support $s_u\in[-\infty,+\infty]$, so all but a vanishing fraction of the prior weight is on arbitrarily large values.  This results in a number of well-known problems, particularly that the value of $s_u$ for any individual who is dominant in every interaction is automatically infinite, which makes it impossible to tell any two such individuals apart.

One solution to these problems is to impose a better prior on~$s_u$ and a natural choice is a logistic prior~\cite{DS73,Whelan17}.  Consider again the quantity $p_0$ defined in Eq.~\eqref{eq:p0}, which is the probability that an individual with strength~$\lambda$ dominates against the average individual:
\begin{equation}
p_0 = {\lambda\over\lambda+1} = {\e^s\over\e^s+1}.
\end{equation}
In the absence of any information to the contrary, we assume this probability to be uniformly distributed between zero and one---the minimally informative or maximum-entropy prior $P(p_0)=1$.  Then the corresponding prior on the score~$s$ is
\begin{equation}
P(s) = P(p_0) {\dd p_0\over\dd s} = {\dd p_0\over\dd s}
     = {1\over(\e^s+1)(\e^{-s}+1)},
\label{eq:logistic}
\end{equation}
which is the logistic distribution.  This is the prior we use in our work.

Now instead of maximizing the likelihood of Eq.~\eqref{eq:likelihood} we maximize the posterior probability, assuming that the prior on $q$ is also uniform.  It is important to recognize that the resulting estimate, like all maximum a posteriori (MAP) estimates, depends on the parametrization of the likelihood, and specifically in this case on whether we maximize over $s_u$ or~$\lambda_u$.  The maximum with respect to one will not in general lie in the same place as the maximum with respect to the other.  In our work we consider the $s_u$ to be the more fundamental set of variables and maximize the posterior distribution
\begin{equation}
P(s,q,\sigma|x) = P(x,\sigma|s,q) {P(s) P(q)\over P(x)}
\end{equation}
with respect to~$s$.  With $P(s)$ as in Eq.~\eqref{eq:logistic} and uniform $P(q)$, and changing variables back to~$\lambda_u$, this gives us
\begin{align}
&P(s,q,\sigma|x) \nonumber\\
  &= \prod_{r=1}^M
  {(\lambda_{u_r}q_{t_r})^{\sigma_r} [\lambda_{v_r}(1-q_{t_r})]^{1-\sigma_r}\over
    \lambda_{u_r}+\lambda_{v_r}} \prod_{u=1}^N {\lambda_u\over(\lambda_u+1)^2}.
\label{eq:likelihood2}
\end{align}
In addition to regularizing the score parameters, the addition of the prior also eliminates the invariance of the model under a uniform shift of the scores and hence eliminates the need to normalize them.

With the addition of the extra term in~\eqref{eq:likelihood2}, the derivation of the algorithm proceeds as before.  Equations~\eqref{eq:estep2} and~\eqref{eq:mstepq} for the quantities~$\pi_r$ and $q_t$ remain unchanged, while the equation for the strengths~$\lambda_u$ now becomes
\begin{equation}
\lambda_i = {1 + \sum_r \bigl[ \pi_r \delta_{u_ri} + (1-\pi_r) \delta_{v_ri}
  \bigr] \over 2/(\lambda_i+1) +  \sum_u A_{iu}/(\lambda_i+\lambda_u)}.
\label{eq:msteplambda2}
\end{equation}
Other than this change, and the omission of the now-unnecessary normalization step, the algorithm is the same as before.

\section{Results}
In this section we demonstrate our approach with example applications of the MAP estimation procedure of Section~\ref{sec:map} to a set of computer-generated test data and to three real-world data sets, an animal dominance hierarchy, a human hierarchy, and an example from competitive team sports.

\subsection{Synthetic tests}
\label{sec:synthetic}
As a first demonstration of our approach we present the results of a set of tests using synthetic (computer-generated) data.  In these tests we generated a large number of random data sets with $N=100$ individuals each and interactions between pairs chosen uniformly at random (a random graph in the language of network theory).  Because the sparsity of the interactions can affect our ability to perform accurate inference, we study cases with both relatively dense interactions ($M=5000$) and sparser ones ($M=1000$).  Tests were also performed with two different choices of the number of interaction types, $T=5$ and 10.  The winners of the interactions were generated using the model of Section~\ref{sec:method} with independent scores~$s_u$ drawn from the logistic distribution of Eq.~\eqref{eq:logistic} and valence probabilities~$q_t$ drawn uniformly in an interval $[q_\textrm{min},q_\textrm{max}]$ for various values of $q_\textrm{min}$ and~$q_\textrm{max}$ as described below.

For each choice of $M$, $q_\textrm{min}$, and~$q_\textrm{max}$ we generated 1000 data sets and analyzed each one by the MAP estimation method of Section~\ref{sec:map} and also by fitting to a traditional unimodal Bradley-Terry model in which all interactions are considered equivalent---the standard practice for calculations of dominance hierarchies for example.  For each analysis we then ranked the fictional participants according to their inferred scores~$s_u$ and computed a Spearman rank correlation between these inferred ranks and the ground-truth ranks implied by the original scores used to generate the data, thereby testing our ability to recover the true ranking of the individuals.  The results are shown in Table~\ref{tab:synthetic}.

\begin{table}
\begin{center}
\begin{tabular}{cr|ccc}
     &     & \multicolumn{3}{c}{Spearman $R^2$ (this paper)/(traditional)} \\
$M$  & $T$ & $0.5\le q_t\le1$ & $0.25\le q_t\le1$ & $0\le q_t\le1$ \\
\hline
5000 & 5   & 0.88/0.83 & 0.83/0.53 & 0.88/0.42 \\
5000 & 10  & 0.89/0.85 & 0.85/0.52 & 0.89/0.29 \\
1000 & 5   & 0.53/0.50 & 0.43/0.24 & 0.54/0.17 \\
1000 & 10  & 0.54/0.52 & 0.41/0.22 & 0.54/0.11 \\
\end{tabular}
\end{center}
\caption{The results of tests of our method on computer-generated data, compared with analyses of the same data using a standard ranking algorithm that assumes all interactions to be equivalent.  All generated examples have $N=100$ individuals and interactions placed between pairs chosen uniformly at random.  Values of the number of interactions~$M$ and number of interaction types~$T$ are as listed and values of $q_t$ are chosen uniformly at random in the intervals shown.  The results are Spearman~$R^2$ values between the inferred ranking of the individuals and the ground-truth ranking calculated from the parameters used to generate the data, averaged over 1000 random instances.  In each entry $x/y$ the first number~$x$ is the result from the method of this paper and the second~$y$ is from the traditional ranking calculation.  Standard errors are less than $\pm1$ in the final digit in all cases.}
\label{tab:synthetic}
\end{table}

We explore three different choices for the valence probabilities~$q_t$.  In the first column of results in Table~\ref{tab:synthetic} we choose values of~$q_t$ uniformly at random in the interval $[0.5,1]$.  This means that all interactions are dominant in the sense of being instigated or won by the dominant individual more often than not.  In this situation our method does well at recovering the ground-truth ranking with Spearman $R^2>0.5$ in all cases, but the standard Bradley-Terry analysis does almost as well.  This is expected since the standard method assumes that indeed all interactions are dominant.  There is nevertheless some daylight between the two methods---our method does better by a small margin in all cases because it is able to make use of the fact that some interactions are more strongly indicative of dominance than others.

The difference becomes more pronounced in the remaining columns of the table.  In the last column we choose values of~$q_t$ in the interval $[0,1]$, which represents the case where interactions are equally likely to indicate dominance or subordination.  In this case our method has a large advantage, since it can estimate the valence of the interactions from the data while standard methods cannot, and we see that there is indeed a large difference between the results for the two methods in this regime---our method continues to do well at recovering the true ranking but the standard method performs poorly.

The middle column gives results for the intermediate case where $q_t$ falls in the interval $[0.25,1]$, which represents a situation in which most interactions indicate dominance but a few do not.  Again we see in this situation that the method of this paper significantly outperforms the standard analysis because it is able to discern the valence of the interactions.

The effect on the results of the number of interaction types~$T$ is small for both our method and the traditional ranking.  The effect of the number of observations~$M$ on the other hand is quite pronounced: our method returns excellent results for all cases with $M=5000$, no matter the values of the other parameters, but is noticeably poorer for the sparser case of $M=1000$.  The traditional method also performs poorly for $M=1000$, although in most cases it does poorly for $M=5000$ too.

To summarize, our method gives modestly improved rankings in situations where all interactions are indicative of dominance and substantially improved rankings in situations where they are not.  Even in cases where the difference between the methods is small, however, the results may be interesting for other reasons.  In particular, they also give an estimate of the valence probability~$q_t$ for each interaction type, which can be of interest in its own right, as we see in the following sections.

\subsection{Dominance hierarchy in vervet monkeys}
We now turn to applications of our method to real-world data.  Our first example application is to a classic animal dominance hierarchy.  We analyze observations reported by Vilette~\etal~\cite{VBHB20} of 66 wild vervet monkeys in the Samara Private Game Reserve in South Africa between January 2015 and December 2017.  The authors report a total of 11\,664 agonistic encounters between pairs of monkeys, divided into eight types: charge, chase, displace, facial, lunge, physical, supplant, and vocal.  A few other types, such as ``scream'' and ``grab,'' were recorded but were rare---less than 1\% of the total---and were removed from the data for our analysis, as were interactions whose type was unknown.  The remaining data were analyzed using the MAP estimation method of Section~\ref{sec:map}.

\begin{figure}
\begin{center}
\includegraphics[width=8cm]{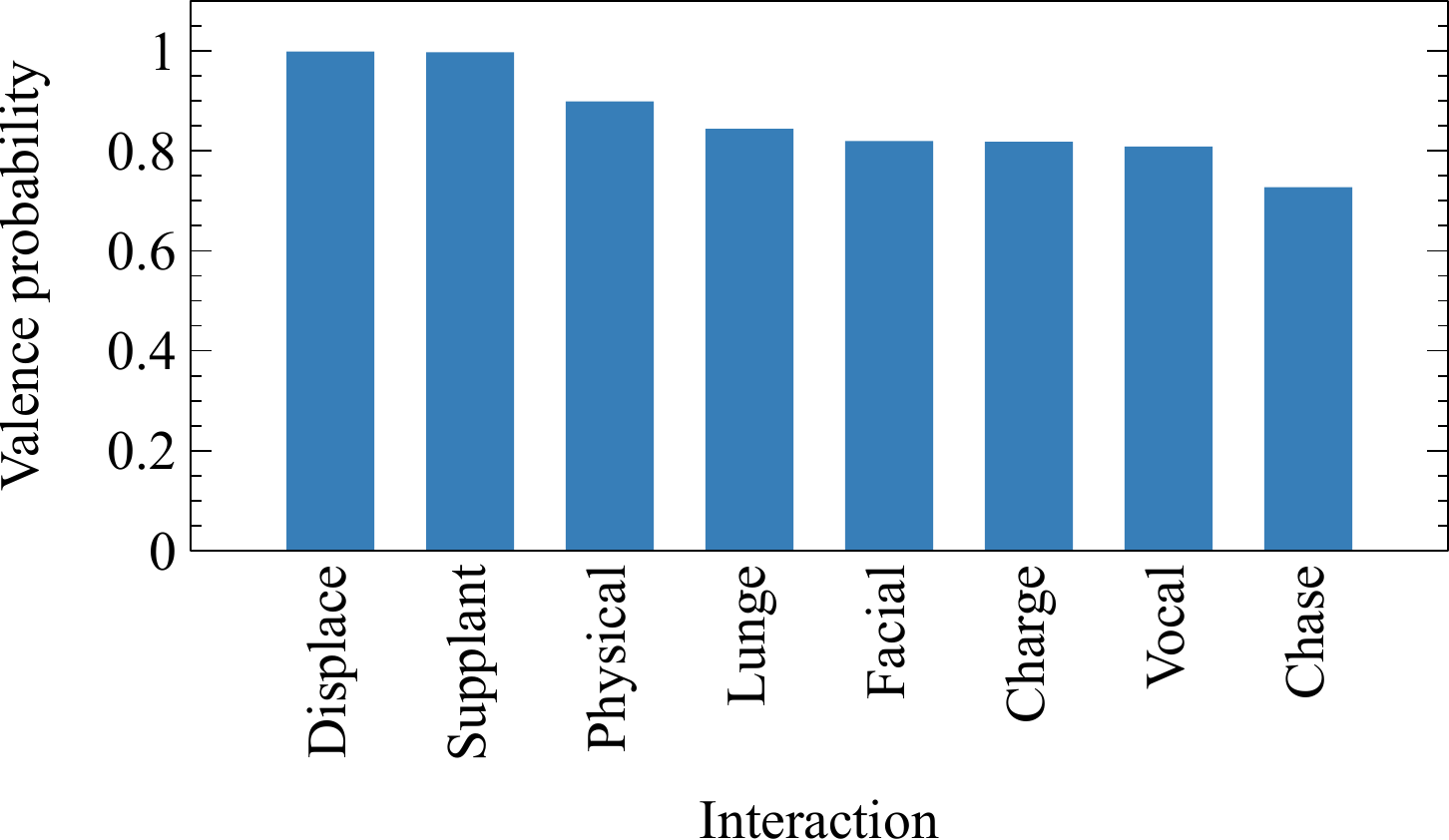}
\end{center}
\caption{The valence probabilities~$q_t$ for the vervet monkey interactions of Vilette~\etal~\cite{VBHB20} as analyzed using the methods of this paper.  These represent the probability that the given interaction will be instigated by a dominant individual, so that values approaching one indicate dominant behaviors and values approaching zero indicate subordinate ones.  In this case all behaviors are dominant on balance, but some more so than others.}
\label{fig:vervetqt}
\end{figure}

Figure~\ref{fig:vervetqt} shows the inferred values of the valence probabilities~$q_t$ for each of the interaction types.  Recall that these values tell us, for each interaction type, the probability that the interaction will be instigated by the dominant individual of a pair.  Values $q_t>\frac12$ indicate interactions that are dominant on average and in this case we find (not surprisingly) that all interaction types are dominant, but that they vary in their degree of dominance.  ``Displace'' and ``supplant,'' for instance, while being arguably the least physical of the interactions, are the most indicative of dominance, while ``chase'' is the least indicative.  ``Chase'' has a valence probability of only $0.728$, implying that about 27\% of the time it is not the chaser but the chasee who is the dominant individual in a pair.  Such interactions are thus less reliable indicators of dominance than ``displace'' and ``supplant.''

\begin{figure}
\begin{center}
\includegraphics[width=8cm]{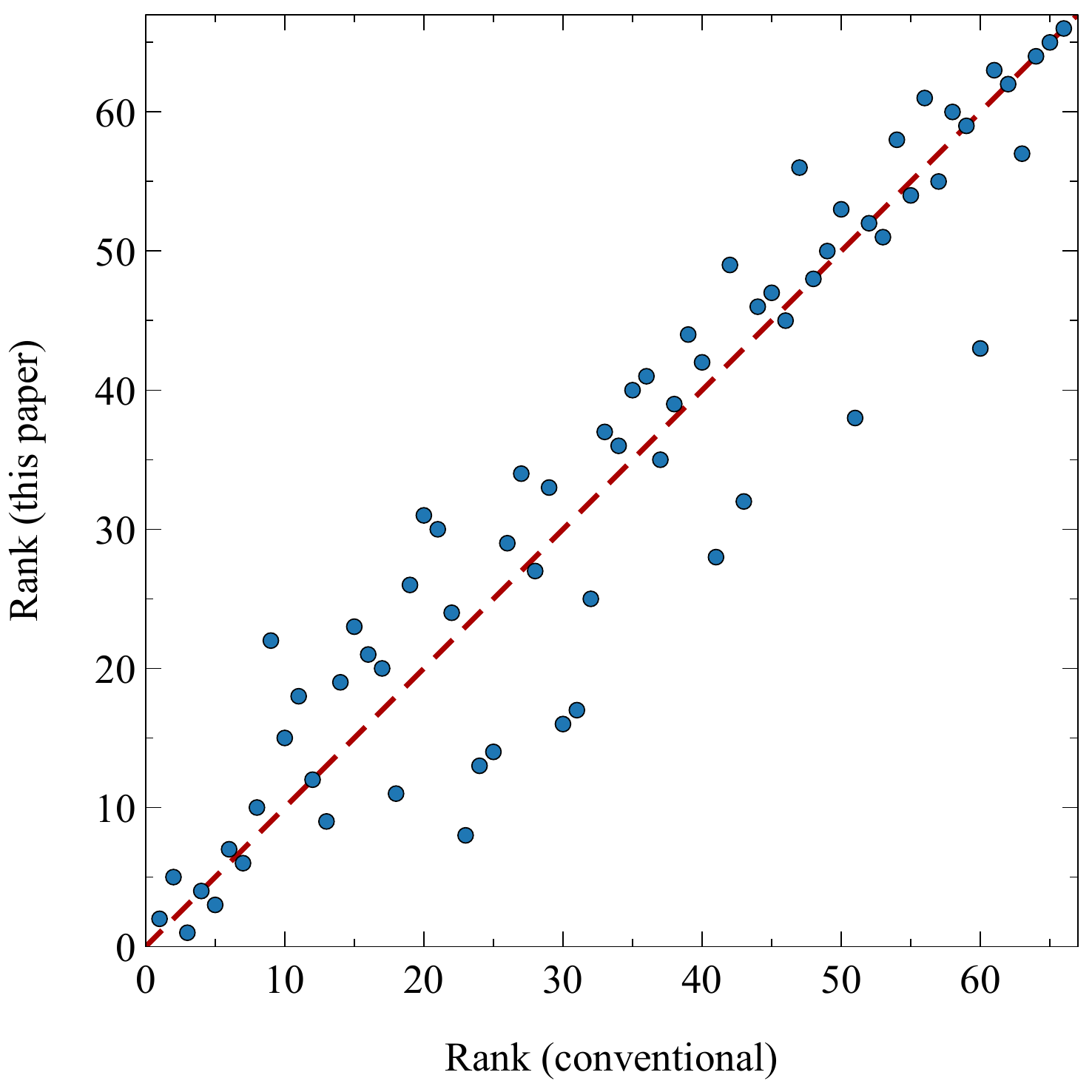}
\end{center}
\caption{Scatter plot of the rankings of the vervet monkeys.  Rankings were computed in two ways: using the method of this paper and by conventional methods that assume all interactions are equally informative (equivalent to assuming that the valence probability~$q_t$ is 1 for all interaction types).  The strength parameters~$\lambda_u$ were computed in both ways then the monkeys were sorted in order of their strengths from highest to lowest to give the rankings plotted here.}
\label{fig:vervetcomp}
\end{figure}

In addition to being informative in their own right, these values now allow us to make a more accurate estimate of the dominance hierarchy: our method automatically weights more (or less) heavily those interactions that strongly (or weakly) indicate dominance.  To illustrate this effect, we show in Fig.~\ref{fig:vervetcomp} a comparison of the rankings of the monkeys, from 1~to~66, computed first as above and second assuming that all interactions are equivalent and equally indicative of dominance.  If the two sets of rankings agreed perfectly, all points in the figure would lie on the dashed line, but, as the figure shows, there are some significant differences between them, in both directions---monkeys who are ranked both higher and lower by our method, up to 17 places different in the most extreme case.

\subsection{Student social network}
\begin{figure}
\begin{center}
\includegraphics[width=8cm]{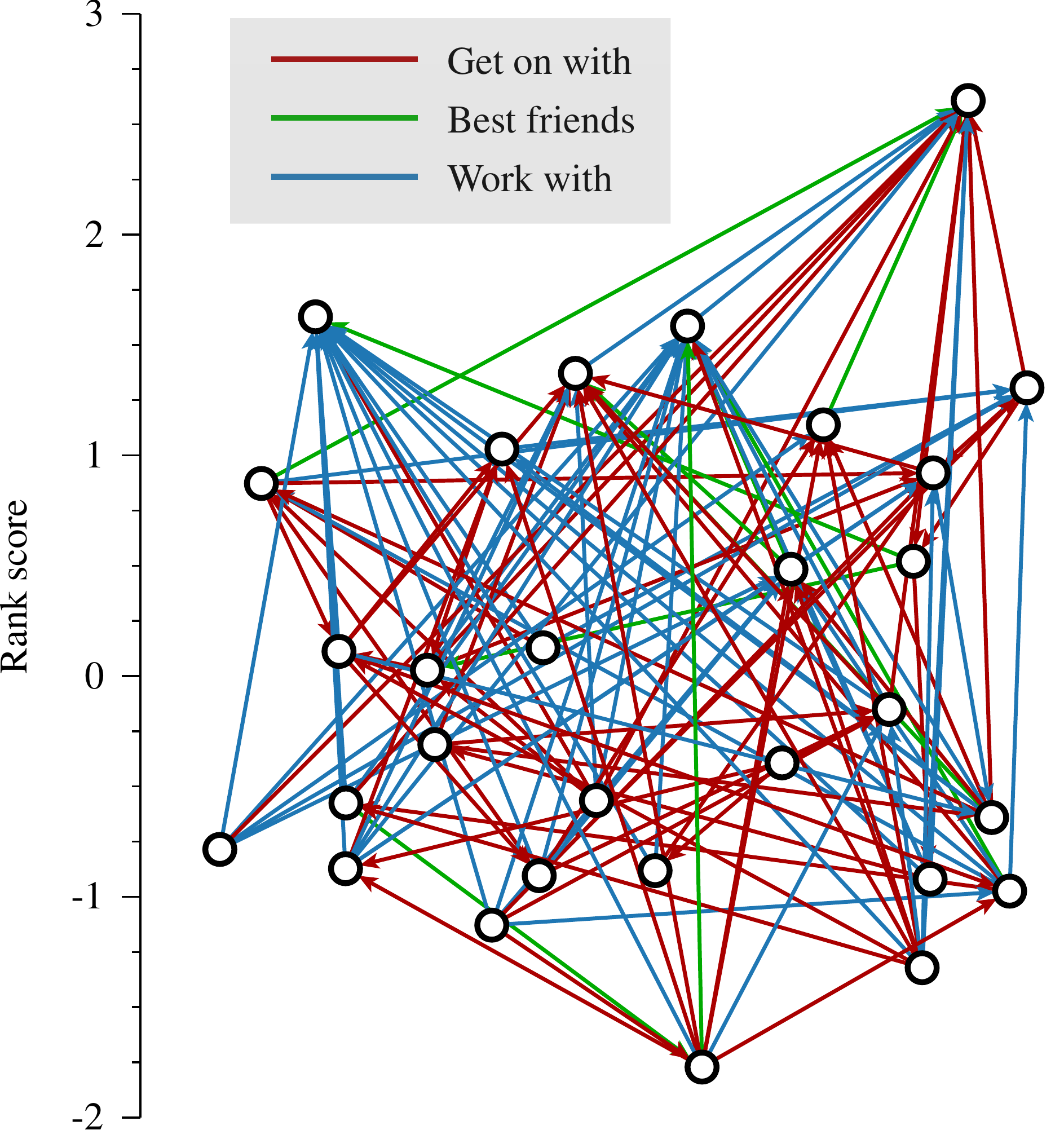}
\end{center}
\caption{Network of 7th grade students in the study of Vickers and Chan~\cite{VC81}.  Nodes represent the students, edges denote the three types of relations as indicated, and the vertical position of each node represents its calculated rank score~$s_u$ on the scale shown at the left.  Following~\cite{BN13} we show only the directed connections between nodes for clarity and we show at most one edge between each pair of nodes.  Edges between nodes connected in both directions are omitted.}
\label{fig:vickersnet}
\end{figure}

For our second example we analyze social interactions among the students in a 7th grade class in Victoria, Australia, as compiled by Vickers and Chan~\cite{VC81}.  In this study the authors interviewed 29 students in a single school class and asked them three questions: (1)~who do you get on with in the class, (2)~who are your best friends in the class, and~(3)~who do you prefer to work with?  The answers to these three questions define three different types of directed relations between the students.  These particular relations are not intrinsically competitive and hence we might not necessarily expect them to define a hierarchy.  However, in other work on friendship among school students it has been found that claims of friendship define a clear hierarchy~\cite{HK88,BN13}, and the same turns out to be true in the present case: using the methods described in this paper we find a strong hierarchy hidden in the data of Vickers and Chan, as shown in Fig.~\ref{fig:vickersnet}.  In this network visualization the nodes represent the students, the directed connections indicate answers to the survey questions, and the vertical position of each node on the page indicates the rank score~$s_u$ assigned it by the analysis, so that nodes higher up the page are more highly ranked.  As we can see, most edges in the network run in an upward direction, meaning that students say they get on with, are friends with, or work with others who are above them in the hierarchy.

On the one hand, this may seem counterintuitive, since one might imagine that friendship and co-working relations should be symmetric: if A is friends with B then surely B is friends with~A as well?  On the other hand, it is common for children (and perhaps adults too) to make or claim ``aspirational'' connections with others: they want to be friends with higher status others~\cite{HK88,DCLV10,BN13}.  These effects lead to asymmetries in reported social networks that can be used, as here, to infer hierarchy.

\begin{figure}
\begin{center}
\includegraphics[width=8cm]{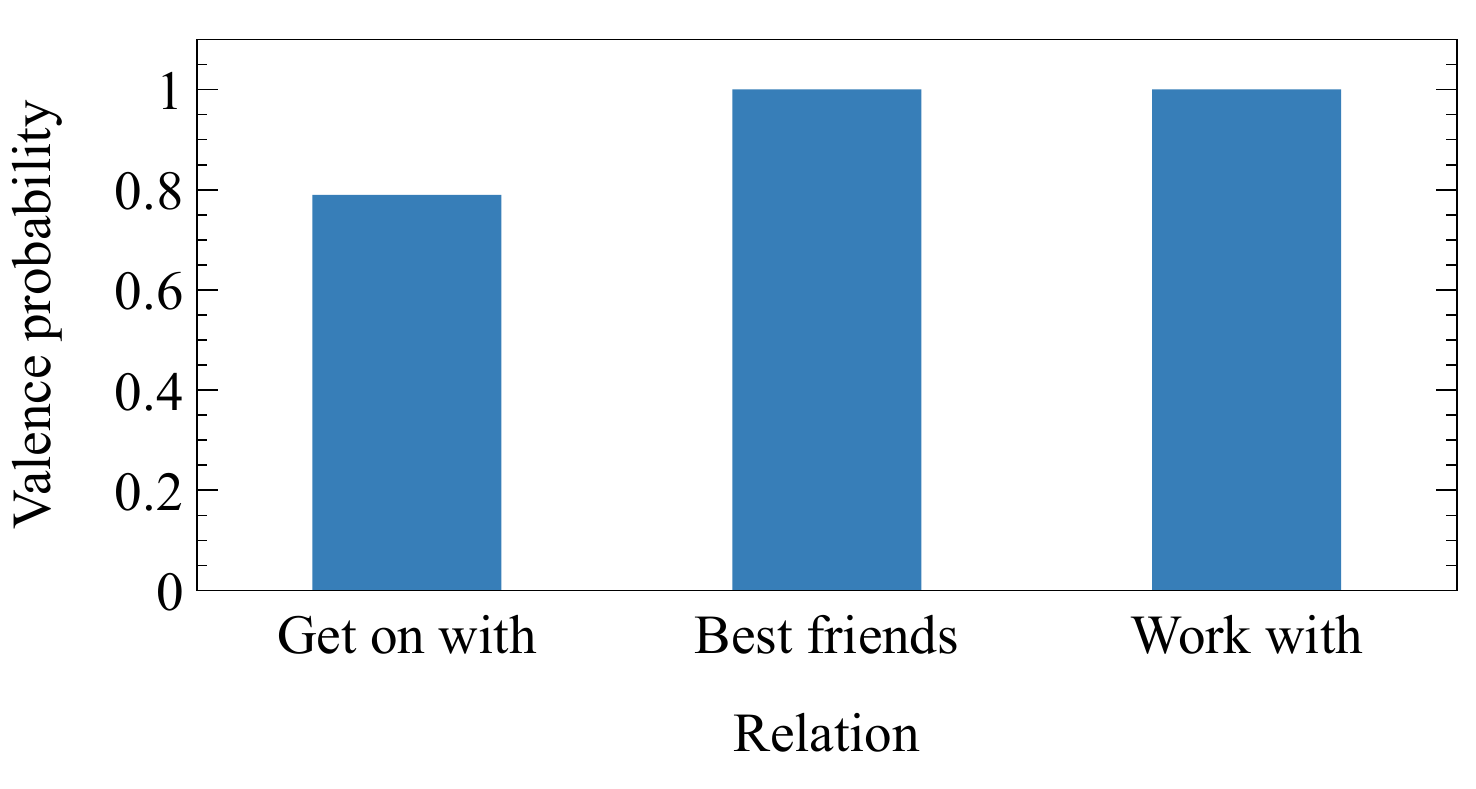}
\end{center}
\caption{Values of the valence probabilities~$q_t$ for the three types of interactions in the student data of Vickers and Chan~\cite{VC81}.}
\label{fig:vickersqt}
\end{figure}

Figure~\ref{fig:vickersqt} shows the values of the parameters~$q_t$ for the three types of connection in this case and here we also find something interesting.  The first type (``who do you get on with?'') is relatively weakly indicative of hierarchy, but for both the second and third types (``best friends'' and ``work with'') the most probable value of $q_t$ is~1, implying that these types of connections are maximally indicative of hierarchy.  Moreover, because the value of $q_t$ is the same for these types the algorithm treats them in an identical way, which means effectively that there are only two types of connections in these data.  The ``best friends'' and ``work with'' interactions are flattened into a single combined interaction type by the analysis.  We have encountered similar flattening of the data in some other examples we have analyzed, including animal hierarchies, a network of tech industry workers, and some online social networks.

\subsection{American football competition}
\label{sec:nfl}
Ranking methods are often applied in sports and athletic competition to rank players or teams.  As a demonstration of this type of application we apply our method to professional American football.  American football provides an interesting example because, unlike association football, it consists of discrete ``plays,'' in which one team (the offense) has possession of the ball and tries to advance it up the field against the other team (the defense).  There are different types of plays teams use to do this, including running plays (the ball is carried by a runner), passing plays (the ball is thrown), and punts (the ball is kicked).  Here we analyze these individual plays as interactions between the teams and compute a ranking of teams based on the pattern of interactions in all (regular season) games in a given playing season.

A key aspect of this analysis is that we use no information about the actual success of the plays---whether they advance the ball, for instance, or whether any points are scored.  Moreover, we specifically remove from the data compulsory plays such as kickoffs and conversions that implicitly signal point scoring, so the only information available to the algorithm is \emph{which} types of plays the teams chose to run.  (We do include field goals, which score points, because these are optional and hence are revealing from a strategic point of view.)

\begin{figure}
\begin{center}
\includegraphics[width=8cm]{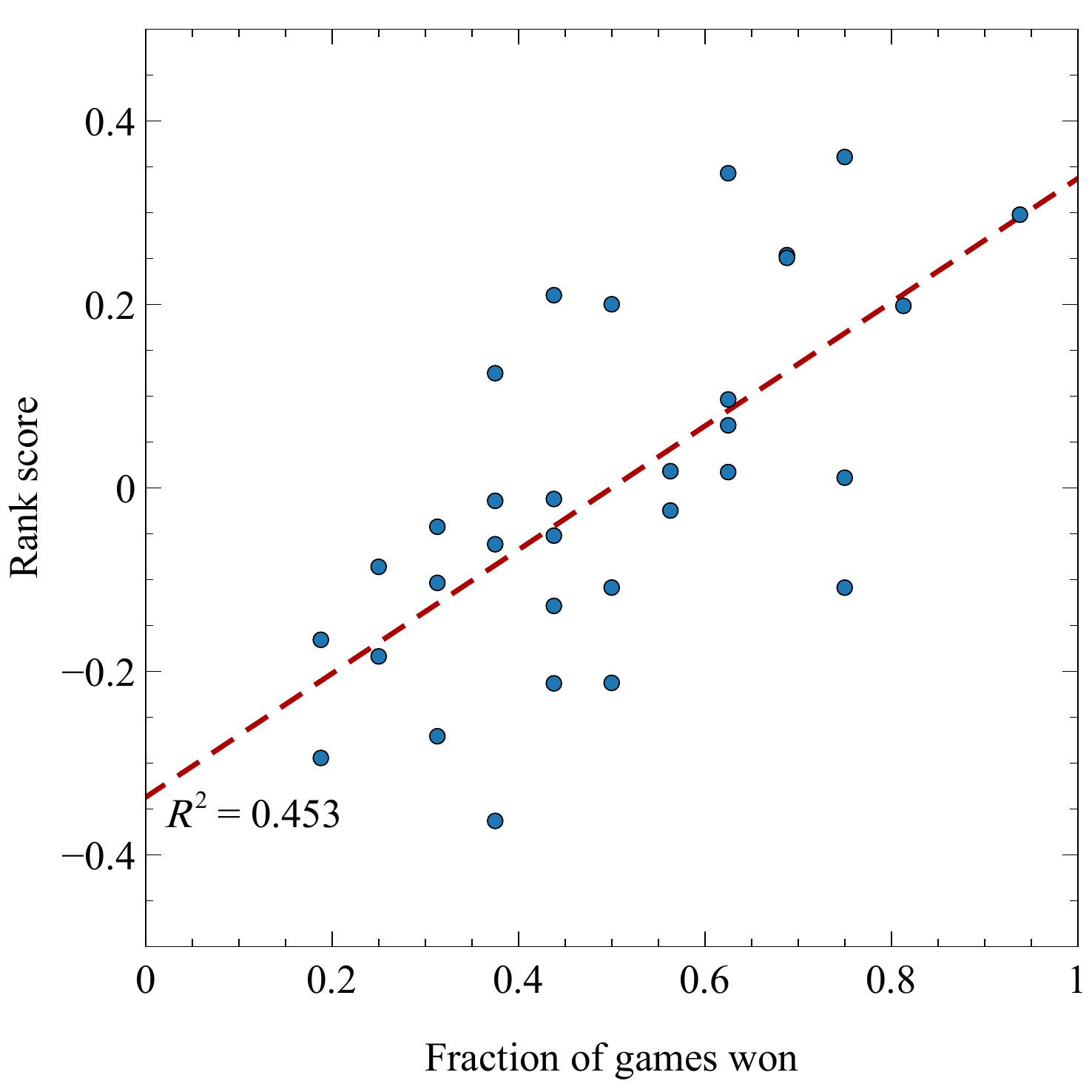}
\end{center}
\caption{The rank scores~$s_u$ of the 32 American football teams in the analysis described here for the 2015 NFL season, plotted against the fraction of games that they won in the same season.}
\label{fig:nfl}
\end{figure}

Even without explicit indicators of success, however, we can extract a meaningful ranking of the teams.  Figure~\ref{fig:nfl} shows an example for the 2015 season of the US National Football League using data from Yurko~\etal~\cite{YVH19}.  With $36\,030$ interactions in total, this example is the largest in this paper, but our algorithm nonetheless runs quickly.  Total running time for the calculation was 11~seconds on a standard laptop computer (\textit{circa} 2022).  The figure shows the inferred rank score~$s_u$ of each of the 32 teams in the league plotted against their actual success during the season, represented by the fraction of games they won.  Though the correlation between the two measures is not perfect, it is substantial and significant ($R^2 = 0.453$, $p<0.0001$).  Note that we should not expect perfect correlation even if our rankings were perfectly accurate, since it is an important aspect of commercially successful spectator sports that they contain an element of randomness.  If the higher ranked team always won there would be little suspense about the outcome of a game and correspondingly little motivation to watch, so the existence of any method that could reliably predict game winners would be a clear sign of an unsuccessful sport (which American football certainly is not).

One might suppose our success at ranking the teams to be a result of the simple fact that winning teams run more plays than losing teams, because they are in possession of the ball more often, but this is not the case.  In fact there is hardly any difference between the number of plays run by the best and worst teams: the top ten teams in our ranking for the 2015 season, for instance, ran an average of 1134 plays in total during the regular season while the bottom ten actually ran a slightly larger number of 1169.

\begin{figure}
\begin{center}
\includegraphics[width=8cm]{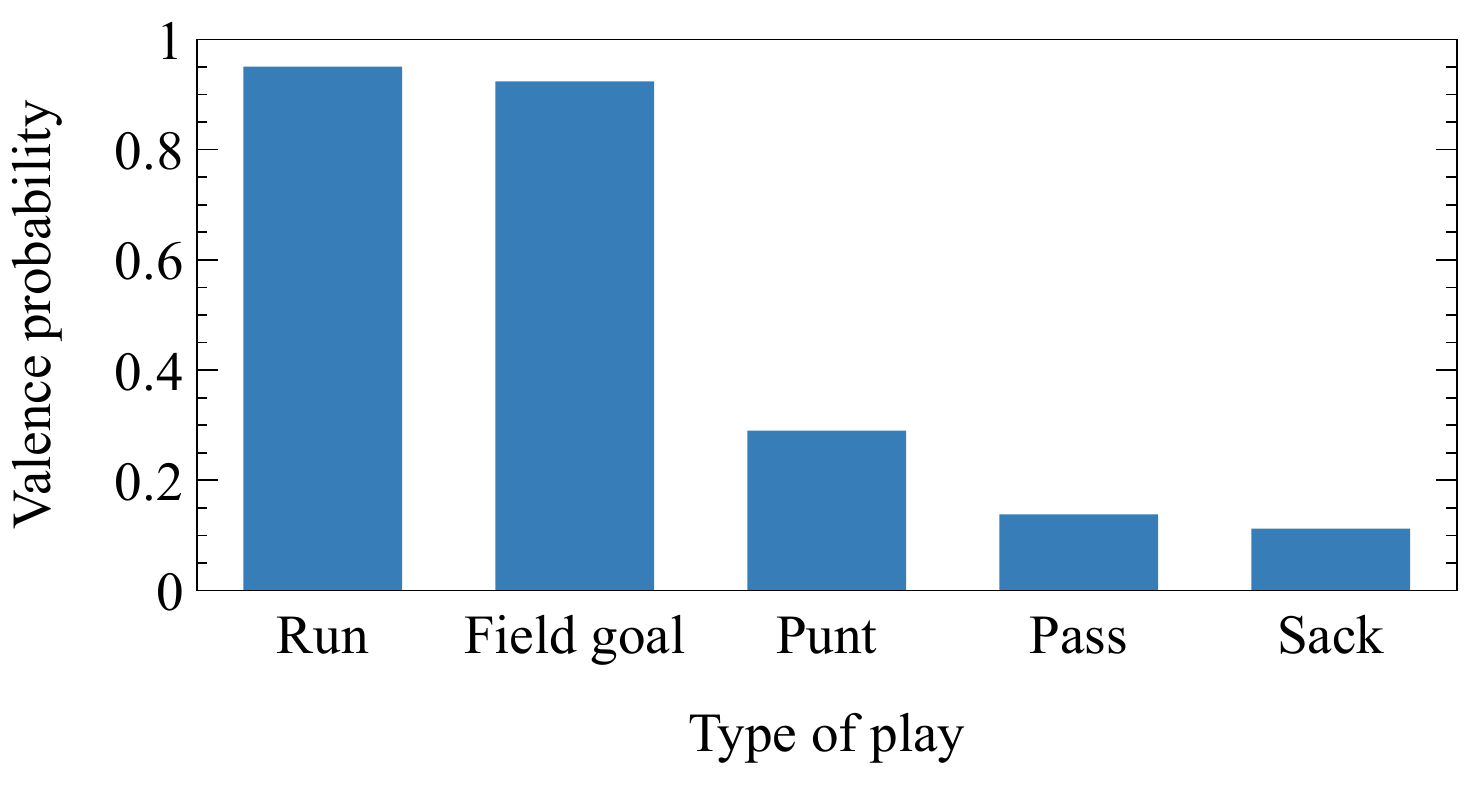}
\end{center}
\caption{The valence probabilities~$q_t$ for each of the five play types used in the ranking of football teams.}
\label{fig:nflqt}
\end{figure}

The ranking of teams is signaled not by the number of plays, but by which plays the teams run.  Figure~\ref{fig:nflqt} shows the values of the valence probabilities~$q_t$ for each of the five play types included in our analysis.  By contrast with our previous examples, not all types of plays signal dominance.  Two types do clearly signal dominance---running plays and field goals---with values of $q_t$ well above~$\frac12$.  The remaining three, however, signal subordination.  Of these, the punt is only used to get rid of the ball when a team knows they are likely to lose it anyway, and hence is a naturally subordinate trait.  And a sack---the player with the ball gets tackled before they can move it up the field---is a clear sign of team weakness.  More surprising is that passing plays, where the ball is thrown, are also a clear sign of weakness.  In general passing plays are some of the most spectacular and successful plays in American football, so one might ask why they are indicative of subordination.  The answer may be that passing plays are challenging to execute and often fail, because for instance the thrown ball is not caught or is intercepted by the opposing team.  This means that weaker teams have to make more attempts to achieve successful passing plays than stronger teams and hence, on balance, passing plays are indicative of weakness.  For instance, during the 2015 season the top ten teams in our ranking made an average of 504 passing plays each while the bottom ten made an average of 620.

\section{Conclusions}
In this paper we have considered the problem of ranking a set of individuals or teams based on pairwise comparisons when there are multiple types of comparison.  Examples include animal dominance hierarchies in which animals use a range of different behaviors to establish or signal dominance, and sports competitions in which teams use a range of different moves or plays against their opponents with varying degrees of success.  We have shown that even if one does not know in advance either the ranking of the individuals or what information each type of interaction conveys, it is possible to infer both from observed interactions.  We have described an efficient method for doing this which combines an expectation-maximization algorithm with a variant of the Bradley-Terry model.

We have presented a number of example applications of the method, including applications to animal and human dominance hierarchies, and an application to the sport of American football.  The method provides a way to sensitively infer rankings taking all interaction types into account and weighting each one appropriately given the information it contains.  At the same time the results shed light on the interactions themselves, telling us, without need for other input, whether each type of interaction is indicative of dominance or subordination, and to what extent.

Natural extensions of the work reported here include the exploration of alternative models for multimodal comparisons, including generalizations of popular models for the unimodal case such as Thurstonian models~\cite{Thurstone27,David88} or models that allow for dependencies between observations~\cite{Cattelan12}.  One could also consider goodness of fit measures to assess the success of our model or any other, model selection to choose between alternatives, or more elaborate inference procedures for the current model, including fully Bayesian approaches similar to those applied in the unimodal case~\cite{DS73,CD12}, which would have the advantage of making the ranking independent of the parametrization of the model.

One difficulty encountered while conducting this work was that, although there is much to be learned from analyses of multimodal competition data, there is, it turns out, relatively little such data currently available.  For instance, although many animals do use multiple modalities to signal or compete for dominance, published studies usually combine these modalities into a single interaction type, making the (potentially questionable) assumption that they are all equally informative.  We encourage researchers working in this area to publish full data sets, including details of the types of interactions, which will allow more nuanced analyses to be conducted.

\begin{acknowledgments}
The author thanks Elizabeth Bruch, Carrie Ferrario, Liza Levina, and Ambuj Tewari for useful conversations.  This work was funded in part by the US National Science Foundation under grant DMS--2005899.  Computer code implementing the method of this paper, along with example data, is available at \verb|https://doi.org/10.1098/rspa.2022.0517| in the supplementary materials.  The data on vervet monkeys, seventh grade students, and American football are all previously published and publicly available.
\end{acknowledgments}

\end{document}